\documentclass[conference]{IEEEtran}

\usepackage[noadjust]{cite}
\usepackage{bbm}
\usepackage{hyperref}
\usepackage{caption}
\usepackage{color}
\usepackage{xspace}
\usepackage[utf8x]{inputenc}
\usepackage{graphicx}
\usepackage[lined, ruled, boxed, commentsnumbered]{algorithm2e}
\usepackage[cmex10]{amsmath}
\usepackage{amssymb}
\usepackage{booktabs}
\usepackage{verbatim}
\usepackage{wrapfig}
\DeclareCaptionType{copyrightbox}
\usepackage{subcaption}
\usepackage{booktabs}
\usepackage{times}
\usepackage{microtype}

\begin{document}
%
\title{Item Recommendation with Evolving User Preferences and Experience\IEEEauthorrefmark{1}}

\newcommand{\squishlist}{
   \begin{list}{$\bullet$}
    { \setlength{\itemsep}{0pt}      \setlength{\parsep}{3pt}
      \setlength{\topsep}{3pt}       \setlength{\partopsep}{0pt}
      \setlength{\leftmargin}{1.5em} \setlength{\labelwidth}{1em}
      \setlength{\labelsep}{0.5em} } }
\newcommand{\squishlisttwo}{
   \begin{list}{$\bullet$}
    { \setlength{\itemsep}{0pt}    \setlength{\parsep}{1pt}
      \setlength{\topsep}{1pt}     \setlength{\partopsep}{0pt}
      \setlength{\leftmargin}{1em} \setlength{\labelwidth}{0.5em}
      \setlength{\labelsep}{0.5em} } }

\newcommand{\squishend}{
    \end{list} 
}


\author{\IEEEauthorblockN{Subhabrata Mukherjee\IEEEauthorrefmark{2},
Hemank Lamba\IEEEauthorrefmark{3} and
Gerhard Weikum\IEEEauthorrefmark{2}
\IEEEauthorblockA{\IEEEauthorrefmark{2}Max Planck Institute for Informatics, \IEEEauthorrefmark{3}Carnegie Mellon University\\
Email: \{smukherjee, weikum\}@mpi-inf.mpg.de, hlamba@cs.cmu.edu}
}}


%

\maketitle

\begin{abstract}
Current recommender systems exploit user and item similarities by collaborative filtering. Some advanced methods also consider
the temporal evolution of item ratings as a global background process.
However, all prior methods disregard the \emph{individual evolution} of a user's {\em experience} level and how
this is expressed in the user's {\em writing} in a review community.

In this paper\let\thefootnote\relax\footnotetext{\IEEEauthorrefmark{1}This is an extended version of the paper published in ICDM 2015~\cite{mukherjee2015jertm}. Refer to~\cite{DBLP:conf/kdd/MukherjeeGW16} for a general (continuous) version of this model with fine-grained temporal evolution of user experience, and resulting language model using Geometric Brownian Motion and Brownian Motion, respectively.}, we model the {\em joint evolution} of {\em user experience},
interest in specific {\em item facets}, {\em writing style}, and {\em rating behavior}.
This way we can generate individual recommendations that
take into account the user's maturity level (e.g., recommending
art movies rather than blockbusters for a cinematography expert).
As only item ratings and review texts are observables, we capture the user's
experience and interests in a {\em latent model} learned from
her reviews, vocabulary and writing style.

We develop a generative HMM-LDA model to trace user
evolution, where the Hidden Markov Model (HMM) traces her 
latent experience progressing over time --- with solely user reviews and ratings as observables over {\em time}.
The facets of a user's interest are drawn from a Latent Dirichlet Allocation (LDA) model derived from her reviews,
as a function of her (again latent) experience level.
In experiments with five real-world datasets, we show
that our model improves the rating prediction over state-of-the-art baselines, by 
a substantial margin.
We also show, in a use-case study, that our model performs
well in the assessment of user experience levels.

\end{abstract}

\IEEEpeerreviewmaketitle

\section{Introduction}

\noindent {\bf Motivation and State-of-the-Art:}
Collaborative filtering algorithms are at the heart of recommender systems
for items like movies, cameras, restaurants and beer.
Most of these methods exploit user-user and item-item similarities
in addition to the history of user-item ratings --- similarities being
based on latent factor models over user and item features~\cite{koren2011advances}, and more recently
on explicit links and interactions
among users~\cite{GuhaWWW2004}\cite{West-etal:2014}.

All these data evolve over {\em time}
leading to bursts in item popularity and other phenomena like anomalies\cite{Gunnemann2014}.
State-of-the-art recommender systems capture these temporal aspects
by introducing global bias components that reflect the evolution of
the user and community as a whole\cite{KorenKDD2010}.
A few models also consider changes in the social neighborhood
of users\cite{MaWSDM2011}.
What is missing in all these approaches, though, is the awareness of
how {\em experience} and {\em maturity} levels evolve in {\em individual users}.

Individual experience is crucial in how users appreciate items, and thus
react to recommendations. For example, a mature cinematographer
would appreciate tips on art movies much more than
recommendations for new blockbusters. Also, the facets of an item that a user focuses on change with experience. For example, 
a mature user pays more attention to narrative, light effects, and style
rather than actors or special effects.
Similar observations hold for ratings of wine, beer, food, etc.

Our approach
advances state-of-the-art by tapping review texts,
modeling their properties as latent factors,
using them to
explain and predict item ratings as a function of a user's experience evolving over time.
Prior works considering review texts (e.g., \cite{mcauleyRecSys2013, wang2011, mukherjeeSDM2014, lakkarajuSDM2011, WangKDD2011})
did this only to learn topic similarities in a
static, snapshot-oriented manner, without considering time at all. The only prior work~\cite{mcauleyWWW2013}, considering time, ignores the text of user-contributed reviews in harnessing their experience. However, 
user experience and their interest in specific item facets
at different timepoints can often be observed only {\em indirectly}
through their ratings, and more {\em vividly} through her vocabulary and writing style in reviews.

\noindent {\bf Use-cases:}
Consider the reviews and ratings by two users on a ``Canon DSLR'' camera about the facet camera {\em lens}.
\vspace{-0.15em}
\squishlisttwo
\item {\small \em User 1: My first DSLR. Excellent camera, takes great pictures in HD, without a doubt it brings honor to its name. [Rating: 5]
}
\item {\small \em User 2: The EF 75-300 mm lens is only good to be used outside. The 2.2X HD lens can only be used for specific items; filters are useless if ISO, AP,... are correct. The short 18-55mm lens is cheap and should have a hood to keep light off lens. [Rating: 3]
}
\squishend
\vspace{-0.15em}
\noindent The second user is clearly more experienced than the first one,
and more reserved about the lens quality of that camera model.
Future recommendations for the second user should take into consideration the user's maturity.
As a second use-case, consider the following reviews of Christopher Nolan movies where the facet of interest is the non-linear {\em narrative style}.
\squishlisttwo
\item {\small \em User 1 on Memento (2001): ``Backwards told is thriller noir-art empty ultimately but compelling and intriguing this.''}
\item {\small \em User 2 on The Dark Knight (2008): ``Memento was very complicated. The Dark Knight was flawless. Heath Ledger rocks!''}
\item {\small \em User 3 on Inception (2010): ``Inception is a triumph of style over substance. It is complex only in a structural way, not in terms of plot. It doesn't unravel in the way Memento does.''}
\squishend
\vspace{-0.15em}
\noindent The first user does not appreciate complex narratives, making
fun of it by writing her review backwards. The second user prefers simpler
blockbusters. The third user seems to appreciate the complex narration style
of Inception and, more of, Memento. We would consider this maturity level of the more experienced User 3 to generate future recommendations to her.


\noindent{\bf Approach:} We model the joint evolution of {\em user experience}, interests in specific {\em item facets}, {\em writing style}, and {\em rating behavior} in a community. As only item ratings and review texts are directly observed, we capture a user's experience and interests by a latent model learned from her reviews, and vocabulary. 
All this is conditioned on {\em time}, considering the {\em maturing rate} of
a user. Intuitively, a user gains experience not only by writing many reviews, but she also needs to continuously improve the quality of her reviews. This varies for different users, as some enter the community being experienced.
This allows us to generate individual recommendations that
take into account the user's maturity level and interest in specific facets
of items, at different timepoints.

We develop a generative HMM-LDA model for a user's
evolution, where the Hidden Markov Model (HMM) traces her 
latent experience progressing over time, and the Latent Dirichlet Allocation (LDA) model captures her interests in specific item facets as a function of her (again, latent) experience level.
The only explicit input to our model is the ratings and review texts
upto a certain timepoint; everything else -- especially the user's
experience level -- is a latent variable.
The output is the predicted ratings for the user's reviews following
the given timepoint.
In addition, we can derive interpretations of a user's experience
and interests by salient words in the distributional vectors for
latent dimensions. 
Although it is unsurprising to see users writing sophisticated words with more experience, we observe something more interesting. For instance in  specialized communities like {\tt \small beeradvocate.com} and {\tt \small ratebeer.com}, experienced users write more descriptive and {\em fruity} words to depict the beer taste (cf. Table~\ref{tab:beerTopics}).
Table~\ref{tab:facetWords} shows a snapshot of the words used at different experience levels to depict the facets {\em beer taste}, {\em movie plot} and {\em bad journalism}, respectively.

We apply our model to $12.7$ million ratings from $0.9$ million users on $0.5$ million items in five different communities on movies, food, beer, and news media, achieving an improvement of $5\%$ to $35\%$ for the mean squared
error for rating predictions over several competitive baselines.
We also show that users at the same (latent) experience level 
do indeed exhibit similar vocabulary, and facet interests.
Finally, a use-case study in a news community to identify experienced {\em citizen journalists} demonstrates that our model captures user maturity fairly well.

\noindent{\bf Contributions:} To summarize, this paper introduces the following novel elements:
\vspace{-0.25em}
\squishlisttwo
\item[a)] This is the first work that considers the progression of user experience as expressed through the text of item reviews, thereby elegantly combining text and time.
\item[b)] An approach to capture the natural {\em smooth} temporal progression in user experience factoring in the {\em maturing rate} of the user, as expressed through her writing. 
\item[c)] Offers interpretability by learning the vocabulary usage of users at different levels of experience.
\item[d)] A large-scale experimental study in {\em five} real world datasets from different communities like movies, beer and food; and an interesting use-case study in a news community.
\squishend
\vspace{-0.25em}

\begin{table}
\small
\begin{tabular}{p{1.3cm}p{1.8cm}p{2.1cm}p{2.0cm}}
\toprule
\centering
\bf{Experience} & \bf{Beer} &\bf{Movies} &\bf{News}\\
\midrule
Level 1 & bad, shit & stupid, bizarre & bad, stupid\\
Level 2 & sweet, bitter & storyline, epic & biased, unfair\\
Level 3 & caramel finish, coffee roasted & realism, visceral, nostalgic & opinionated, fallacy, rhetoric\\
\bottomrule
\end{tabular}
\caption{Vocabulary at different experience levels.}
\label{tab:facetWords}
\vspace{-1.5em}
\end{table}

\section{Overview}
\label{sec:overview}

\subsection{Model Dimensions}

Our approach is based on the intuition that there is a strong coupling between the \emph{facet preferences} of a user, her \emph{experience}, \emph{writing style} in reviews, and \emph{rating behavior}.
All of these factors jointly evolve with \emph{time} for a given user.

We model the user experience progression through discrete stages, so a state-transition model is natural. Once this decision is made, a Markovian model is the simplest, and thus natural choice. This is because the experience level of a user at the current instant $t$ depends on her experience level at the previous instant $t$-$1$. As experience levels are latent (not directly observable), a Hidden Markov Model is appropriate. Experience progression of a user depends on the following factors:


\squishlisttwo
\item \emph{Maturing rate} of the user which is modeled by her \emph{activity} in the community. The more engaged a user is in the community, 
the higher are the chances that she gains experience and advances in writing sophisticated reviews,
and develops taste to appreciate specific facets.
\item \emph{Facet preferences} of the user in terms of focusing on particular
facets of an item (e.g., narrative structure rather than special effects).
With increasing maturity, the taste for particular facets becomes more refined.
\item \emph{Writing style} of the user,  as expressed by the language model at her current level of experience. More sophisticated vocabulary and writing style indicates higher probability of progressing to a more mature level.
\item \emph{Time difference} between writing successive reviews. 
It is unlikely for the user's experience level to change from that of her last review in a short time span (within a few hours or days).
\item \emph{Experience level difference}: Since it is unlikely for a user to directly progress to say level $3$  from level $1$ without passing through level $2$, the model at each instant decides whether the user should stay at current level $l$, or progress to $l$+$1$.
%
\squishend

In order to learn the \emph{facet preferences} and \emph{language model} of a user at different levels of experience, we use \emph{Latent Dirichlet Allocation} (LDA). In this work, we assume each review to
refer to exactly one item. Therefore, the facet distribution of items is 
expressed in the facet distribution of the review documents.

We make the following assumptions for the generative process of writing a review by a user at time $t$ at  experience level $e_t$:

\squishlisttwo
\item A user has a distribution over \emph{facets},
where the facet preferences of the user depend on her experience level $e_t$.
\item A facet has a distribution over \emph{words} where the words used to describe a facet depend on the user's vocabulary at experience level $e_t$.
Table \ref{tab:amazonTopics} shows salient words for two facets of
Amazon movie reviews at different levels of user experience, automatically extracted by our latent model.
The facets are latent, but we can interpret them as {\em plot/script} and
{\em narrative style}, respectively.
\squishend

\begin{table}
\small
\begin{tabular}{p{8.7cm}}
\toprule
\textbf{Level 1:} stupid people supposed wouldnt pass bizarre totally cant\\
\textbf{Level 2:}storyline acting time problems evil great times didnt money ended simply falls pretty\\
\textbf{Level 3:} movie plot good young epic rock tale believable acting\\
\textbf{Level 4:} script direction years amount fast primary attractive sense talent multiple demonstrates establish\\
\textbf{Level 5:} realism moments filmmaker visual perfect memorable recommended genius finish details defined talented visceral nostalgia\\
\midrule
\textbf{Level 1:} film will happy people back supposed good wouldnt cant\\
\textbf{Level 2:} storyline believable acting time stay laugh entire start funny\\
\textbf{Level 3 \& 4:} narrative cinema resemblance masterpiece crude undeniable admirable renowned seventies unpleasant myth nostalgic\\
\textbf{Level 5:} incisive delirious personages erudite affective dramatis nucleus cinematographic transcendence unerring peerless fevered\\
\bottomrule
\end{tabular}
\caption{Salient words for two facets at
five experience levels in movie reviews.}
\label{tab:amazonTopics}
         \vspace{-1.5em}
\end{table}

\noindent As a sanity check for our assumption of the coupling between user {\em experience}, {\em rating behavior}, {\em language} and {\em facet preferences}, we perform experimental studies reported next.

\label{sec:study}

\subsection{Hypotheses and Initial Studies}

\noindent{\bf Hypothesis 1: Writing Style Depends on Experience Level.}

We expect users at different  experience levels to have divergent Language Models (LM's) --- with experienced users having a more sophisticated writing style and vocabulary than amateurs.
To test this hypothesis, we performed initial studies over two popular communities\footnote{Data available at http://snap.stanford.edu/data/}:
1) BeerAdvocate ({\small \tt beeradvocate.com}) with $1.5$ million reviews from $33,000$ users
and 2) Amazon movie reviews ({\small \tt amazon.com}) with $8$ million reviews from $760,000$ users.
Both of these span a period of about 10 years.

In BeerAdvocate, a user gets \emph{points} on the basis of  likes received for her reviews, ratings from other 
users, number of posts written, diversity and number of beers rated, time in the community, etc. 
We use this points measure as a proxy for the user's \emph{experience}. 
In Amazon, reviews get \emph{helpfulness} votes from other users.  For {each user},
we aggregate these votes over all her reviews and take this as a proxy for her experience.

\begin{figure}[h]
         \vspace{-1em}
        \centering
        \begin{subfigure}[b]{0.25\textwidth}
                \includegraphics[width=\textwidth]{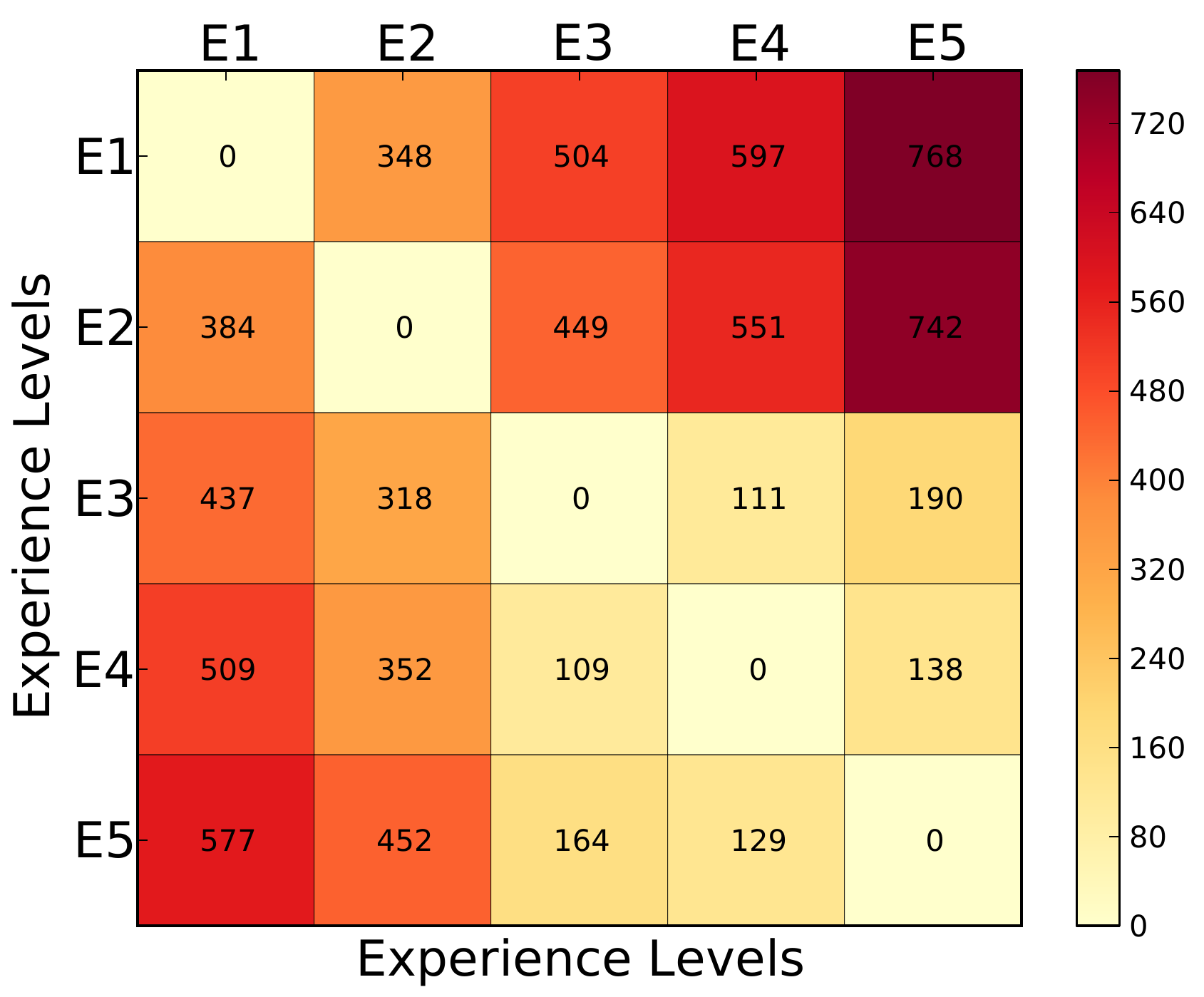}
 \caption{Divergence of language model\\ as a function of experience.}
 \label{fig:expertiseLangDivergence}
        \end{subfigure}%
        ~\hfill
        \begin{subfigure}[b]{0.25\textwidth}
                \includegraphics[width=\textwidth]{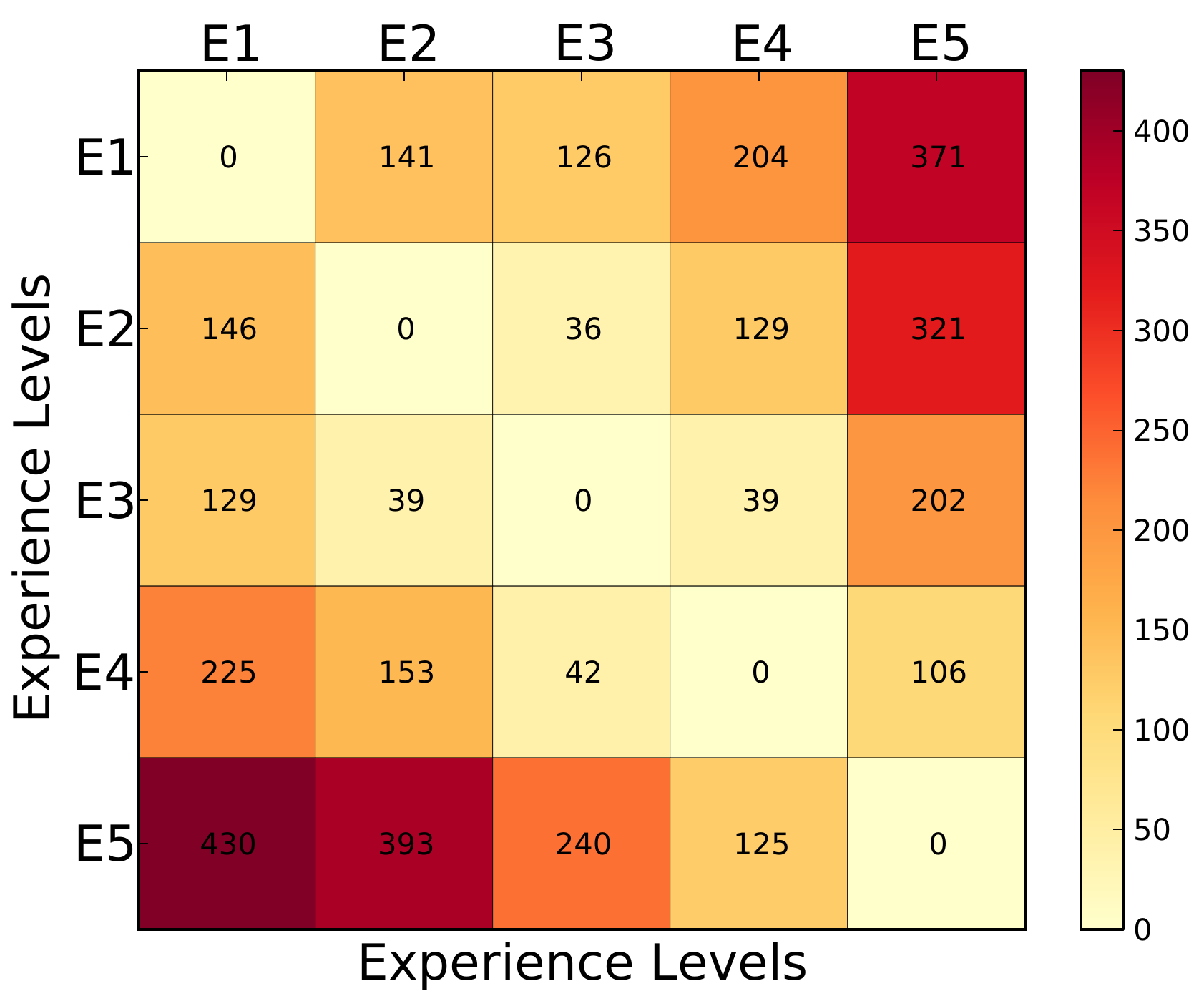}
 \caption{Divergence of facet preference\\ as a function of experience.}
 \label{fig:expertiseFacetDivergence}
        \end{subfigure}
         \caption{$KL$ Divergence as a function of experience.}
         \label{fig:expertiseDivergence}
         \vspace{-1em}
\end{figure}

We partition the users into $5$ bins, based on the points / helpfulness votes received, each representing one of the experience levels. For each bin, we aggregate the review texts of all users in that bin and construct a unigram language model. 
The heatmap of Figure  \ref{fig:expertiseLangDivergence} shows the \emph{Kullback-Leibler} (KL) divergence between the LM's of different experience levels, for the BeerAdvocate case. 
The Amazon reviews lead to a very similar heatmap, which is omitted here.
The main observation is that the KL divergence is higher --- the larger the difference is between the experience
levels of two users. This confirms our hypothesis about the coupling of experience and user language.
%

\noindent{\bf Hypothesis 2: Facet Preferences Depend on\\ Experience Level.}

The second hypothesis underlying our work is that users at similar levels of experience have similar facet preferences.
In contrast to the LM's where words are {\em observed}, facets are {\em latent} so that validating or
falsifying the second hypothesis
is not straightforward.  We performed a three-step study:
\squishlisttwo
\item We use Latent Dirichlet Allocation (LDA)~\cite{Blei2003LDA} to compute
a latent facet distribution $\langle f_k \rangle$ of \emph{each review}. 
\item We run Support Vector Regression (SVR)~\cite{drucker97} for \emph{each user}. 
The user's item rating in a review is the response variable, with the facet proportions in the review given 
by LDA as features.
 The regression weight $w^{u_e}_k$ is then interpreted as the preference of user $u_e$ for facet $f_k$. 
\item Finally, we aggregate these facet preferences for each experience level $e$ to get the corresponding facet preference distribution given by $\ \ <\frac{\sum_{u_e} exp(w^{u_e}_k)}{\#u_e}>$.
\squishend

Figure~\ref{fig:expertiseFacetDivergence} shows the $KL$ divergence between the facet preferences of users at different experience levels in BeerAdvocate. 
We see that the divergence clearly increases with the difference in user experience levels;
this confirms the hypothesis.
The heatmap for Amazon is similar and omitted. 

{\em Note} that Figure~\ref{fig:expertiseDivergence} shows how a {\em change} in the experience level can be detected. This is not meant to predict the experience level, which is done by the model in Section~\ref{sec:inference}.



\section{Building Blocks of our Model}

Our model, presented in the next section, builds on and compares itself
against various baseline models as follows.

\subsection{Latent-Factor Recommendation}

According to the standard latent factor model (LFM)~\cite{korenKDD2008}, the rating assigned by a user $u$ to an item $i$ is given by:
\begin{equation}
\label{eq.1}
 rec(u, i) = \beta_g + \beta_u + \beta_i + \langle \alpha_u, \phi_i \rangle
\end{equation}

where $\langle .,. \rangle$ denotes a scalar product. $\beta_g$ is the average rating of all items by all users. $\beta_u$ is the offset of the average rating given by user $u$  from the global rating.
Likewise $\beta_i$ is the rating bias for item $i$. 
$\alpha_u$ and $\phi_i$ are the latent factors associated with user $u$ and item $i$, respectively. These latent factors are learned using gradient descent by minimizing the mean squared error ($MSE$) between observed ratings $r(u,i)$ and predicted ratings $rec(u,i)$:
$MSE = \frac{1}{|U|} \sum_{u,i \in U} (r(u,i) - rec(u, i))^2$

\subsection{Experience-based Latent-Factor Recommendation}

The most relevant baseline for our work is the ``user at learned rate'' model of ~\cite{mcauleyWWW2013},
which exploits that users at the same experience level have similar rating behavior even if their ratings are temporarily far apart. 
Experience of each user $u$ for item $i$ is modeled as a latent variable $e_{u,i} \in \{1...E\}$.
Different recommenders are learned for different experience levels. Therefore Equation~\ref{eq.1} is parameterized as:
\begin{equation}
\label{eq.3}
 rec_{e_{u,i}}(u, i) = \beta_g(e_{u,i}) + \beta_u(e_{u,i}) + \beta_i(e_{u,i}) + \langle \alpha_u(e_{u,i}), \phi_i(e_{u,i}) \rangle
\end{equation}

The parameters are learned using Limited Memory BFGS
with the additional constraint that experience levels should be non-decreasing over the reviews written by a user over time. 

However, this is significantly different from our approach. All of these models work on the basis of only user {\em rating behavior}, and ignore the review texts completely. Additionally, the {\em smoothness} in the evolution of parameters between experience levels is enforced via $L_2$ regularization, and does not model the {\em natural} user maturing rate (via HMM) as in our model. Also note that in the above parametrization, an experience level is estimated for each user-item pair. However, it is rare that a user reviews the same item multiple times. In our approach, we instead trace the evolution of users, and not user-item pairs. 


\subsection{User-Facet Model}

In order to find the facets of interest to a user, \cite{rosenzviUAI2004} extends Latent Dirichlet Allocation (LDA) to include authorship information. 
Each document $d$ is considered to have a distribution over authors. 
We consider the special case where each document has exactly one author
$u$ associated with a Multinomial distribution $\theta_u$ over facets $Z$ with a symmetric Dirichlet prior $\alpha$. The facets have a Multinomial distribution $\phi_z$ over words $W$ drawn from a vocabulary $V$ with a symmetric Dirichlet prior $\beta$. 
Exact inference is not possible due to the intractable coupling between $\Theta$ and $\Phi$. Two ways for approximate inference are 
MCMC techniques like Collapsed Gibbs Sampling and Variational Inference.
The latter is typically much more complex and computationally expensive.
In our work, we thus use sampling.

\subsection{Supervised User-Facet Model}

The generative process described above is unsupervised and does not take the ratings in reviews into account. Supervision is difficult to build into MCMC sampling where ratings are continuous values, as in communities like {\small\tt newstrust.net}.
For discrete ratings, a review-specific Multinomial rating distribution $\pi_{d,r}$ can be learned as in~\cite{linCIKM2009, ramageKDD2011}. 
Discretizing the continuous ratings into buckets bypasses the problem to some extent, but results in loss of information. Other approaches~\cite{lakkarajuSDM2011, mcauleyRecSys2013, mukherjeeSDM2014} overcome this problem by learning the feature weights separately from the user-facet model.

\begin{figure}
  \vspace{-0.5em}
\centering
 \includegraphics[scale=0.3]{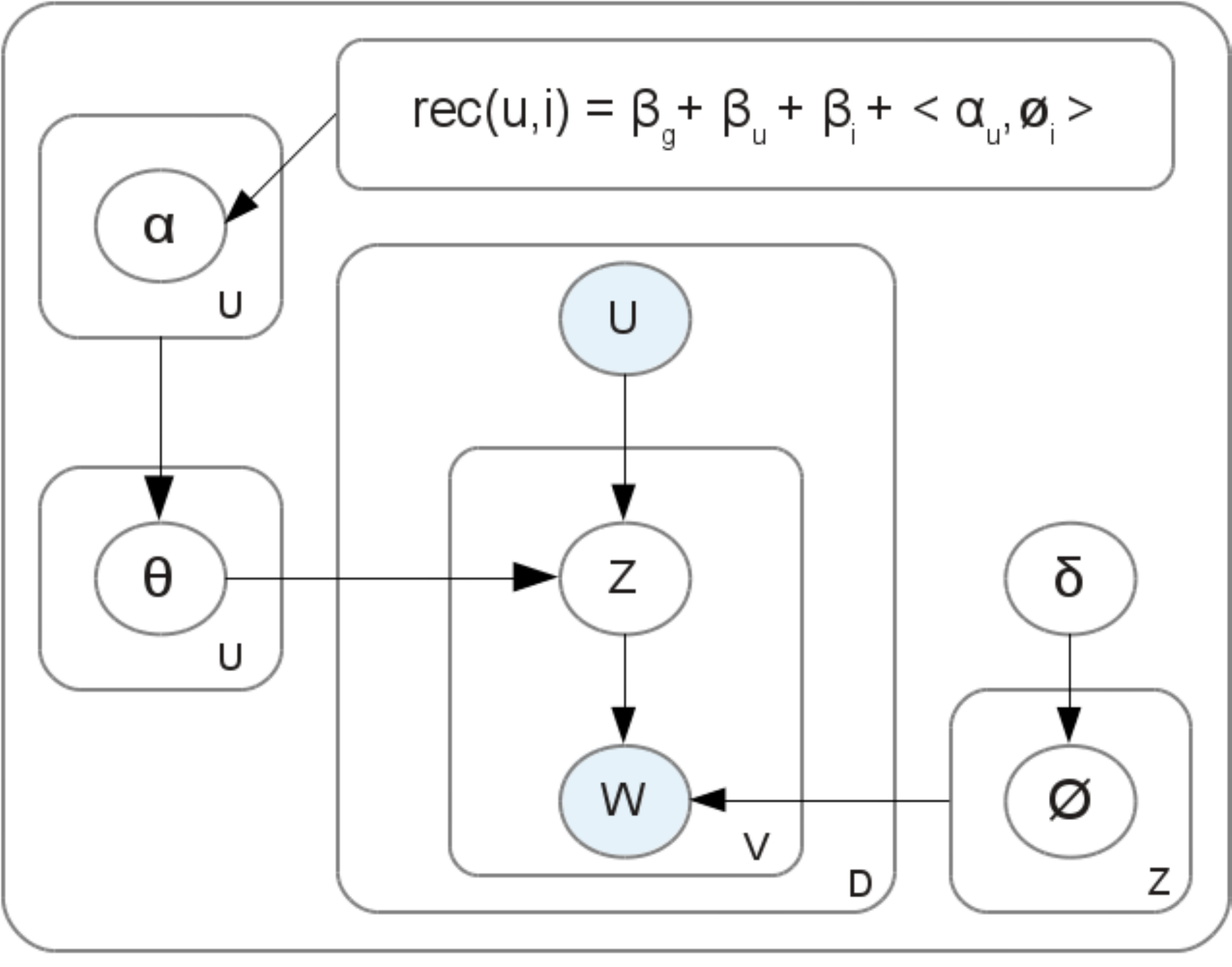}
  \caption{\small Supervised model for user facets and ratings.}
  \label{fig:1}
  \vspace{-1em}
\end{figure}
 
An elegant approach using Multinomial-Dirichlet Regression is proposed in~\cite{mimnoUAI2008} to incorporate arbitrary types of observed continuous or categorical features. Each facet $z$ is associated with a vector $\lambda_z$ whose dimension equals the number of features. Assuming $x_d$ is the feature vector for document $d$, the Dirichlet hyper-parameter $\alpha$ for the document-facet Multinomial distribution $\Theta$ is parametrized as $\alpha_{d,z}=exp(x_d^T\lambda_z)$. The model is trained using stochastic \emph{EM} which alternates between 1) sampling facet assignments from the posterior distribution conditioned on words and features, and 2) optimizing $\lambda$ given the facet assignments using L-BFGS.
Our approach, explained in the next section, follows a similar approach to couple 
the User-Facet Model and the Latent-Factor Recommendation Model (depicted in Figure~\ref{fig:1}).


\section{Joint Model: User Experience, \\ Facet Preference, Writing Style}
\label{sec:inference}


\begin{figure}
 \includegraphics[scale=0.35]{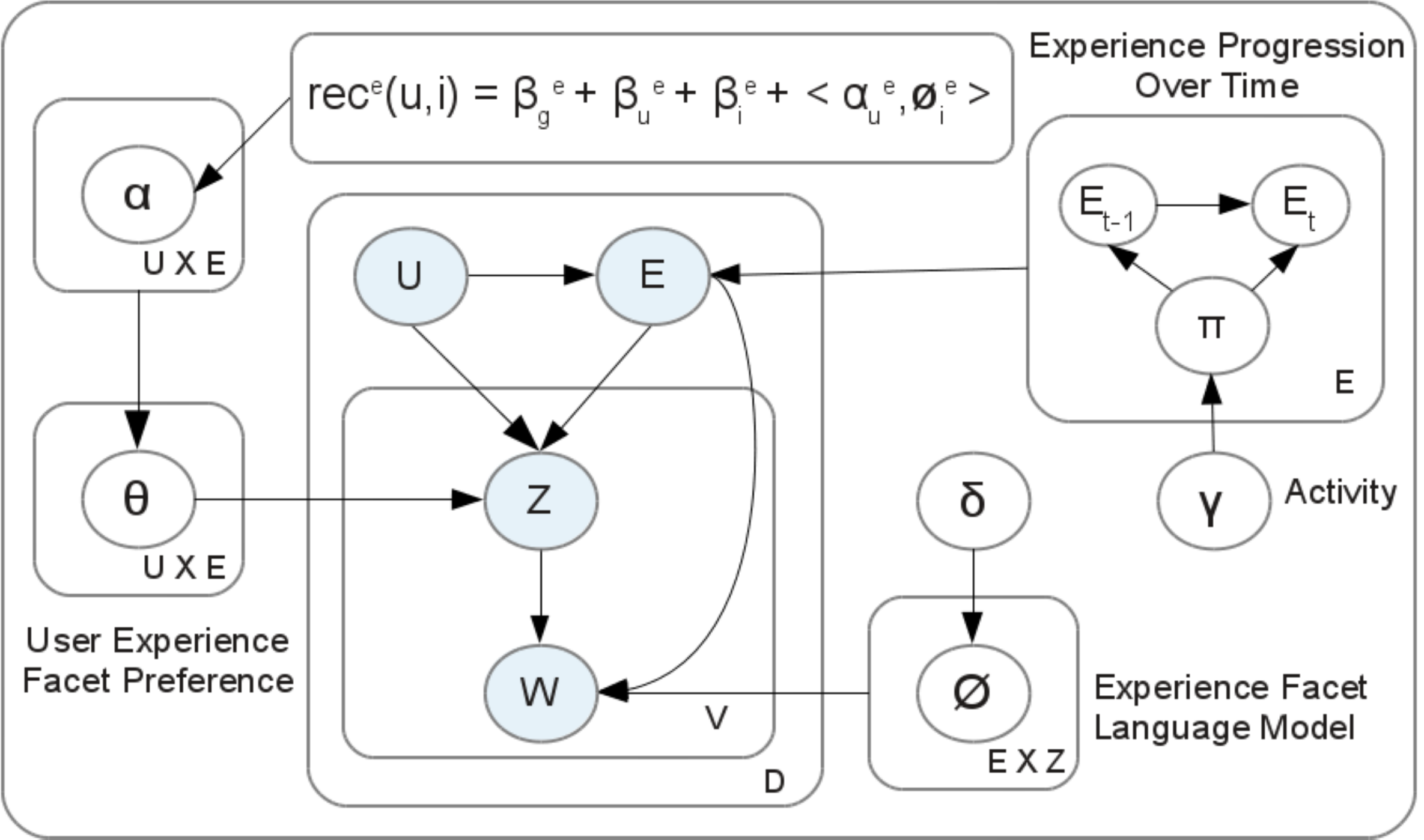}
  \caption{\small Supervised model for user experience, facets, and ratings.}
  \label{fig:2}
  \vspace{-1.5em}
 \end{figure}

We start with a \emph{User-Facet Model (UFM)} (aka. Author-Topic Model~\cite{rosenzviUAI2004}) based on \emph{Latent Dirichlet Allocation} (LDA), where {users} have a distribution over {facets} and facets have a distribution over {words}. This is to determine the facets of interest to a user. These facet preferences can be interpreted as latent item factors in the traditional \emph{Latent-Factor Recommendation Model} (LFM)~\cite{korenKDD2008}. However, the LFM is supervised as opposed to the UFM. 
It is not obvious how to incorporate supervision into the UFM to predict ratings. 
The user-provided ratings of items can take continuous values (in some review communities), 
so we cannot incorporate them into a UFM with a Multinomial distribution of ratings.
We propose an \emph{Expectation-Maximization (EM)} approach to incorporate supervision, where the
latent facets are estimated in an \emph{E-Step} using \emph{Gibbs Sampling}, and \emph{Support Vector Regression} (SVR)~\cite{drucker97} is used in the \emph{M-Step} to learn the feature weights and predict ratings. Subsequently, we incorporate a layer for \emph{experience} in the UFM-LFM model, where the experience levels are drawn from a \emph{Hidden Markov Model} (HMM) in the \emph{E-Step}. The experience level transitions depend on the evolution of the user's \emph{maturing rate}, \emph{facet preferences}, and \emph{writing style} over \emph{time}. The entire process is a supervised generative process of generating a review based on the experience level of a user hinged on our HMM-LDA model.


\subsection{Generative Process for a Review}

Consider a corpus with a set $D$ of review documents denoted by $\{d_1 \dots d_D\}$. For \emph{each user}, all her documents are ordered by timestamps $t$ when she wrote them, such that $t_{d_i}<t_{d_j}$ for $i<j$. Each document $d$ has a sequence of $N_d$ words denoted by $d=w_1 \dots .w_{N_d}$. 
Each word is drawn from a vocabulary $V$ having unique words indexed by $\{1 \dots V\}$. 
Consider a set of $U$ users involved in writing the documents in the corpus, 
where $u_d$ is the author of document $d$. 
Consider an ordered set of experience levels  $\{e_1,e_2,...e_E\}$ 
where each  $e_i$ is from a set $E$,
and a set of facets $\{z_1,z_2,...z_Z\}$ where each $z_i$ is from a set $Z$ of possible facets. 
Each document $d$ is associated with a rating $r$ and an item $i$.

At the time $t_d$ of writing the review $d$, the user $u_d$ has experience level $e_{t_d} \in E$. We assume that her experience level transitions follow a distribution $\Pi$ with a Markovian assumption 
and certain constraints.
This means the experience level of $u_d$ at time $t_d$ depends on her 
experience level when writing the previous document at time $t_{d-1}$. 

$\pi_{e_i}(e_j)$ denotes the probability of progressing to experience level $e_j$ from experience level $e_i$, with the constraint $e_j \in \{e_i, e_i+1\}$. This means at each instant the user can either stay at her current experience level, or move to the next one.

%
The experience-level transition probabilities
depend on the \emph{rating behavior}, \emph{facet preferences}, and \emph{writing style} of the user. 
The progression also takes into account the 1) \emph{maturing rate} of  $u_d$ modeled by the intensity of her activity in the community,
and 2) the \emph{time gaps} between writing consecutive reviews.
We incorporate these aspects in 
a prior
for the user's transition rates,
$\gamma^{u_d}$, defined as:
\[
 \gamma^{u_d} = \frac{D_{u_d}}{D_{u_d} + D_{avg}} + {\lambda (t_d - t_{d-1})}
\]

$D_{u_d}$ and $D_{avg}$ denote the number of reviews written by ${u_d}$ and the average number of reviews per user in the community, respectively.
Therefore the first term
models the user activity with respect to the community average. The second term reflects the time difference between successive reviews.
The user experience is unlikely to change from the level when writing the previous review just a few hours or days ago.
$\lambda$ controls the effect of this time difference, and is set to a very small value. Note that if the user writes very infrequently, the second term may go up. But the first term which plays the \emph{dominating} role in this prior will be very small with respect to the community average in an active community, bringing down the influence of the entire prior.
Note that the constructed HMM encapsulates all the factors for experience progression outlined in Section~\ref{sec:overview}.

At experience level $e_{t_d}$, user $u_d$ has a Multinomial facet-preference distribution $\theta_{u_d, e_{t_d}}$.
From this distribution she draws a facet of interest $z_{d_i}$ for the $i^{th}$ word in her document.
For example, a user at a high level of experience may choose to write on the beer ``hoppiness'' or ``story perplexity'' in a movie.
The word that she writes depends on the facet chosen and the language model for 
her current experience level. Thus, she draws a word from the multinomial distribution $\phi_{e_{t_d}, z_{d_i}}$ with a symmetric Dirichlet prior $\delta$. 
For example, if the facet chosen is beer {\em taste} or movie {\em plot}, an experienced user may choose to use the words ``coffee roasted vanilla'' and ``visceral'', whereas an inexperienced user may use ``bitter'' and ``emotional'' resp.

Algorithm~\ref{algo.3} describes this generative process for the review; Figure~\ref{fig:2}
depicts it visually in plate notation for graphical models.
We use \emph{MCMC} sampling for inference on this model.

\begin{algorithm}[h]
\small
\label{algo.3}
\SetAlgoLined
\DontPrintSemicolon
\For {each facet $z = 1, ... Z$ and experience level $e = 1, ... E$} {choose $\phi_{e, z} \sim Dirichlet(\beta)$} \;
\For {each review $d = 1, ... D$} {
  Given user $u_d$ and timestamp $t_d$\;
  /*Current experience level depends on previous level*/ \;
  1. Conditioned on $u_d$ and previous experience $e_{t_{d-1}}$, 
  choose $e_{t_d} \sim \pi_{e_{t_{d-1}}}$ \;
  /*User's facet preferences at current experience level
are influenced by supervision via $\alpha$ $~$ -- $~$ scaled by
hyper-parameter $\rho$ controlling influence of supervision*/\;
  2. Conditioned on supervised facet preference $\alpha_{u_d, e_{t_d}}$ 
of $u_d$ at experience level $e_{t_d}$ scaled by $\rho$, choose $\theta_{u_d, e_{t_d}} \sim  Dirichlet(\rho \times \alpha_{u_d, e_{t_d}})$ \;
  \For {each word $i = 1, ... N_d$} {
    /*Facet is drawn from user's experience-based facet interests*/ \;
    3. Conditioned on $u_d$ and $e_{t_d}$ choose a facet $z_{d_i} \sim Multinomial(\theta_{u_d, e_{t_d}})$ \;
    /*Word is drawn from chosen facet and user's 
vocabulary at her current experience level*/ \;
    4. Conditioned on $z_{d_i}$ and $e_{t_d}$ choose a word $w_{d_i} \sim Multinomial(\phi_{e_{t_d}, z_{d_i}})$ \;
  }
  /*Rating computed via Support Vector Regression with \\
chosen facet proportions as input features to learn $\alpha$*/ \;
  5. Choose $r_d \sim F(\langle {\alpha_{u_d, e_{t_d}}}, \phi_{e_{t_d}, z_d} \rangle)$ \;
}
\caption{Supervised Generative Model for a User's $~~~~~$ Experience, Facets, and Ratings}
\label{algo.3}
\end{algorithm}


\subsection{Supervision for Rating Prediction}

The latent item factors $\phi_i$ in Equation~\ref{eq.3} correspond to the latent facets $Z$ in Algorithm~\ref{algo.3}.
Assume that we have some estimation of the latent facet distribution $\phi_{e,z}$ of each document after one iteration of MCMC sampling, where $e$ denotes the experience level at which a document is written, and let $z$ denote a latent facet of the document. 
We also have an estimation of the preference of a user $u$ for facet $z$ 
at experience level $e$ given by $\theta_{u,e}(z)$.

For each user $u$, we compute a supervised regression function $F_u$
for the user's numeric ratings 
with the -- currently estimated -- experience-based facet distribution $\phi_{e,z}$ of her reviews as input features
and the ratings as output.

The learned feature weights $\langle\alpha_{u,e}(z)\rangle$ indicate the user's preference for facet $z$ at experience level $e$. These feature weights are used to modify $\theta_{u,e}$ to
 attribute more mass to the facet for which $u$ has a higher preference at level $e$. This is reflected in the next sampling iteration, when we draw a facet $z$ from the user's facet preference distribution $\theta_{u,e}$ smoothed by $\alpha_{u,e}$, and then draw a word from $\phi_{e,z}$. This
sampling process is repeated until convergence.

In any latent facet model, it is difficult to set the hyper-parameters.
Therefore, 
most prior work assume
symmetric Dirichlet priors with heuristically chosen concentration parameters. 
%
Our approach is to {\em learn} the concentration parameter $\alpha$ of a {\em general} (i.e., asymmetric) Dirichlet prior for Multinomial distribution $\Theta$ -- where we optimize these hyper-parameters to learn user ratings for documents at a given experience level.

\subsection{Inference}

We describe the inference algorithm to estimate the distributions $\Theta$, $\Phi$ and $\Pi$
from observed data.
For {each user}, we compute the conditional distribution over the set of hidden variables $E$ and $Z$ for all the words $W$ in a review. 
The exact computation of this distribution is intractable. 
We use {\em Collapsed Gibbs Sampling} \cite{Griffiths02gibbssampling} to estimate the conditional distribution for each hidden variable, which is computed over the current assignment for all other hidden variables, and integrating out other parameters of the model.

Let $U, E, Z$  and $W$ be the set of all users, experience levels, facets and words in the corpus. 
In the following, $i$ indexes a document and $j$ indexes a word in it.

The joint probability distribution is given by:
\begin{equation}
\small
 \begin{aligned}
P(U, E, Z, W, \theta,\phi ,\pi;\alpha ,\delta ,\gamma) =
\prod_{u = 1}^U \prod_{e = 1}^E \prod_{i = 1}^{D_u} \prod_{z = 1}^Z \prod_{j = 1}^{{N_{d_u}}}\{\\
\underbrace{P(\pi_e;\gamma^u ) \times P(e_i|\pi_e)}_\text{\small experience transition distribution} \times \underbrace{P(\theta _{u,e};\alpha_{u,e}) \times P(z_{i,j}|\theta_{u,e_i})}_\text{\small user experience facet distribution}\\
\times \underbrace{P(\phi_{e,z};\delta) \times P(w_{i,j}|\phi_{e_i,z_{i,j}})}_\text{\small experience facet language distribution}\}
\end{aligned}
\end{equation}

Let $n(u, e, d, z, v)$ denote the count of the word $w$
occurring in document $d$ written by user $u$ at experience level $e$ belonging to facet $z$. In the following equation,
 $(.)$ at any position in a distribution indicates summation of the above counts  
for the respective argument.

Exploiting conjugacy of the Multinomial and Dirichlet distributions, 
we can integrate out $\Phi$ from the above distribution to obtain the posterior distribution
$P(Z|U,E; \alpha )$ of the latent variable $Z$ given by:

{\footnotesize
\[\prod_{u=1}^U \prod_{e=1}^E \frac{\Gamma(\sum_{z} \alpha_{u,e,z})\prod_{z} \Gamma(n(u, e, ., z, .)+ \alpha_{u,e,z})}{\prod_{z}{\Gamma(\alpha_{u,e,z})\Gamma(\sum_{z} n(u, e, ., z, .) + \sum_z \alpha_{u,e,z})}}\]
}
where $\Gamma$ denotes the Gamma function.
 
Similarly, by integrating out $\Theta$, $P(W|E,Z; \delta )$ is given by
{\footnotesize
\[\prod_{e=1}^E \prod_{z=1}^Z \frac{\Gamma(\sum_{v} \delta_v)\prod_{v} \Gamma(n(., e, ., z, v)+ \delta_v)}{\prod_{v}{\Gamma(\delta_v)\Gamma(\sum_{v} n(., e, ., z, v) + \sum_v \delta_v)}}\]
}

Let $m_{e_i}^{e_{i - 1}}$
denote the number of transitions from experience level $e_{i-1}$ to $e_i$ over {\em all} users in the community, with the constraint $e_i \in \{e_{i-1}, e_{i-1}+1\}$. Note that we allow self-transitions for staying at the same experience level. The counts capture the relative difficulty in progressing between different experience levels. For example, it may be easier to progress to level $2$ from level $1$ than to level $4$ from level $3$.

%

%
The state transition probability depending on the previous state, factoring in the user-specific activity rate, is given by:

$P(e_i|e_{i - 1}, u, e_{-i}) = \frac{m_{e_i}^{e_{i - 1}} + I(e_{i - 1} = e_i) + \gamma^u}{m_{.}^{e_{i-1}} + I(e_{i - 1} = e_i) + E \gamma^u}$

where $I(.)$ is an indicator function taking the value $1$ when the argument is true, and $0$ otherwise. The subscript $- i$ denotes the value of a variable excluding the data at the $i^{th}$ position. All the {\em counts} of transitions exclude transitions to and from $e_i$, when sampling a value for the current experience level $e_i$ during Gibbs sampling.
The conditional distribution for the experience level transition is given by:
\begin{equation}
\label{eq.6}
 P(E|U,Z,W) \propto P (E|U) \times P(Z|E, U) \times P(W|Z, E)
\end{equation}

Here the first factor models the rate of experience progression factoring in user activity; the second and third factor models the facet-preferences of user, and language model at a specific level of experience respectively. All three factors combined decide whether the user should stay at the current level of experience, or has matured enough to progress to next level.

In Gibbs sampling, the conditional distribution for each hidden variable is computed based on the current assignment of other hidden variables. The values for the latent variables are sampled repeatedly from this conditional distribution until convergence. In our problem setting we have two sets of latent variables corresponding to $E$ and $Z$ respectively.

We perform Collapsed Gibbs Sampling \cite{Griffiths02gibbssampling} in which we first sample a value for the experience level $e_i$ of the user for the current document $i$, keeping all facet assignments $Z$ fixed. In order to do this, we consider two experience levels $e_{i-1}$ and ${e_{i-1}+1}$. For each of these levels, we go through the current document and all the token positions to compute Equation~\ref{eq.6} --- and choose the level having the highest conditional probability.
Thereafter, we sample a new facet for each word $w_{i,j}$
of the document, keeping the currently sampled experience level of the user for the document fixed.

The conditional distributions for Gibbs sampling for the joint update of the latent variables $E$ and $Z$ are given by:

\begin{equation}
\small
\hspace{-1em}
\begin{aligned}
\textbf{\small E-Step 1: } P(e_i = e |e_{i-1}, u_i = u, \{z_{i,j}=z_j \}, \{w_{i,j}=w_j\}, e_{-i})  \propto & \\
 P(e_i|u, e_{i-1}, e_{-i}) \times \prod_j P(z_j|e_i, u, e_{-i}) \times P(w_j|z_j, e_i, e_{-i}) \propto & \\
\frac{m_{e_i}^{e_{i - 1}} + I(e_{i - 1} = e_i) + \gamma ^u}{m_{.}^{e_{i-1}} + I(e_{i - 1} = e_i) + E\gamma^u} \times \\
 \prod_j \frac{n(u, e, ., z_j, .) + \alpha_{u,e,z_j}}{\sum_{z_j} n(u, e, ., z_j, .) + \sum_{z_j} \alpha_{u,e,z_j}} \times \frac{n(., e, ., z_j, w_j) + \delta}{\sum_{w_j} n(., e, ., z_j, w_j) + V\delta}\\
\noindent \textbf{\small E-Step 2:\quad\quad} P(z_j = z|u_d = u, e_d=e, w_j =w, z_{-j}) \propto &\\
\frac{n(u, e, ., z, .) + \alpha_{u,e,z}}{\sum_{z} n(u, e, ., z, .) + \sum_{z} \alpha_{u,e,z}} \times \frac{n(., e, ., z, w) + \delta}{\sum_{w} n(., e, ., z, w) + V\delta}
\end{aligned}
\label{eq.4}
\end{equation}

The proportion of the $z^{th}$ facet in document $d$ with words $\{w_j\}$ written at experience level $e$ is given by:
\[
\footnotesize
\phi_{e,z}(d) = \frac{\sum_{j=1}^{N_d} \phi_{e,z}(w_j)}{N_d}
\]

For each user $u$, we learn a regression model $F_u$ using these facet proportions in each document as features, along with the user and item biases (refer to Equation~\ref{eq.3}), 
with the user's item rating $r_d$ as the response variable. 
Besides the facet distribution of each document, the biases $<\beta_g(e), \beta_u(e), \beta_i(e)>$ also depend on the experience level $e$.


We formulate the function $F_u$ as Support Vector Regression~\cite{drucker97}, which forms the $M$-$Step$ in our problem:
\begin{equation*}
\small
\begin{aligned}
\textbf{M-Step:    } \min_{\alpha_{u,e}} \frac{1}{2}  {\alpha_{u,e}} & ^T{\alpha_{u,e}} + C \times & \\
 \sum_{d=1}^{D_u} (max(0, |r_d - {\alpha_{u,e}}^T<\beta_g(e), & \beta_u(e), \beta_i(e), \phi_{e,z}(d)>| - \epsilon))^2
 \end{aligned}
  \label{eq.5}
\end{equation*}

The total number of parameters learned is $[E \times Z + E \times 3] \times U$.
Our solution may generate a mix of positive and negative real numbered weights. In order to ensure that 
the concentration parameters of the Dirichlet distribution are positive reals, we take $exp(\alpha_{u,e})$. The learned $\alpha$'s are typically very small, whereas the value of $n(u, e, ., z, .)$ in Equation~\ref{eq.4} is very large. Therefore we scale the $\alpha$'s by a hyper-parameter $\rho$ to control the influence of supervision.
$\rho$ is tuned
using a validation set by varying it from $\{10^0, 10^1... 10^5\}$. In the \emph{E-Step} of the next iteration, we choose $\theta_{u,e} \sim Dirichlet(\rho \times \alpha_{u,e})$. We use the LibLinear\footnote{http://www.csie.ntu.edu.tw/~cjlin/liblinear} package for Support Vector Regression.

\section{Experiments}
\label{sec:experiments}

\noindent {\bf Setup:} We perform experiments with data from five communities in different domains:
BeerAdvocate ({\tt \small beeradvocate.com}) and RateBeer ({\tt \small ratebeer.com}) for beer reviews, Amazon ({\tt\small amazon.com}) for movie reviews, Yelp ({\tt \small yelp.com}) for food and restaurant reviews, and NewsTrust ({\tt \small newstrust.net}) for reviews of news media. Table~\ref{tab:statistics} gives the dataset statistics\footnote{http://snap.stanford.edu/data/, http://www.yelp.com/dataset\_challenge/}. 
We have a total of $12.7$ million reviews from $0.9$ million users from all of the five communities combined. The first four communities are used for product reviews, from where we extract the following quintuple for our model $<userId, itemId, timestamp, rating, review>$. 
NewsTrust is a special community, which we discuss in Section~\ref{sec:usecases}.

For all models, we used the three most recent reviews of each user as withheld test data. All experience-based models consider the \emph{last} experience level reached by each user, and corresponding learned parameters for rating prediction. In all the models, we group {\em light} users with less than $50$ reviews in {\em training} data into a background model, treated as a single user, to avoid modeling from sparse observations. We do not ignore any user. During the {\em test} phase for a light user, we take her parameters from the background model. We set $Z=20$ for BeerAdvocate, RateBeer and Yelp facets; and $Z=100$ for Amazon movies and NewsTrust which have much richer latent dimensions. For experience levels, we set $E=5$ for all. However, for NewsTrust and Yelp datasets our model categorizes users to belong to one of {\em three} experience levels.

\begin{table}
\centering
\begin{tabular}{lrrr}
\toprule
\bf{Dataset} & \bf{\#Users} & \bf{\#Items} & \bf{\#Ratings}\\
\midrule
\bf{Beer (BeerAdvocate)} & 33,387 & 66,051 & 1,586,259\\
\bf{Beer (RateBeer)} & 40,213 & 110,419 & 2,924,127\\
\bf{Movies (Amazon)} & 759,899 & 267,320 & 7,911,684\\
\bf{Food (Yelp)} & 45,981 & 11,537 & 229,907\\
\bf{Media (NewsTrust)} & 6,180 & 62,108 & 134,407\\
\midrule
\bf{TOTAL} & 885,660 & 517,435 & 12,786,384\\
\bottomrule
\end{tabular}
\caption{Dataset statistics.}
\label{tab:statistics}
\vspace{-1em}
\end{table}

\subsection{Quantitative Comparison}

\begin{table}
\centering
\begin{tabular}{p{3.5cm}p{0.8cm}p{0.6cm}p{0.6cm}p{0.6cm}p{0.6cm}}
\toprule
\bf{Models} & \bf{Beer} & \bf{Rate} & \bf{News} & \bf{Amazon} & \bf{Yelp}\\
 & \bf{Advocate} & \bf{Beer} & \bf{Trust} & &\\
\midrule
Our model & 0.363 & 0.309 & 0.373 & 1.174 & 1.469\\
(most recent experience level) & & & & &\\
{\bf f)} Our model & 0.375 & 0.362 & 0.470 & 1.200 & 1.642\\
(past experience level) & & & & &\\
{\bf e)} User at learned rate & 0.379 & 0.336 & 0.575 & 1.293 & 1.732\\
{\bf c)} Community at learned rate & 0.383 & 0.334 & 0.656 & 1.203 & 1.534\\
{\bf b)} Community at uniform rate & 0.391 & 0.347 & 0.767 & 1.203 & 1.526\\
{\bf d)} User at uniform rate & 0.394 & 0.349 & 0.744 & 1.206 & 1.613\\
{\bf a)} Latent factor model & 0.409 & 0.377 & 0.847 & 1.248 & 1.560\\
\bottomrule
\end{tabular}
\caption{MSE comparison of our model versus baselines.}
\label{fig:MSE}
\vspace{-1.5em}
\end{table}

\begin{figure}
\vspace{-1.5em}
\centering
 \includegraphics[scale=0.45]{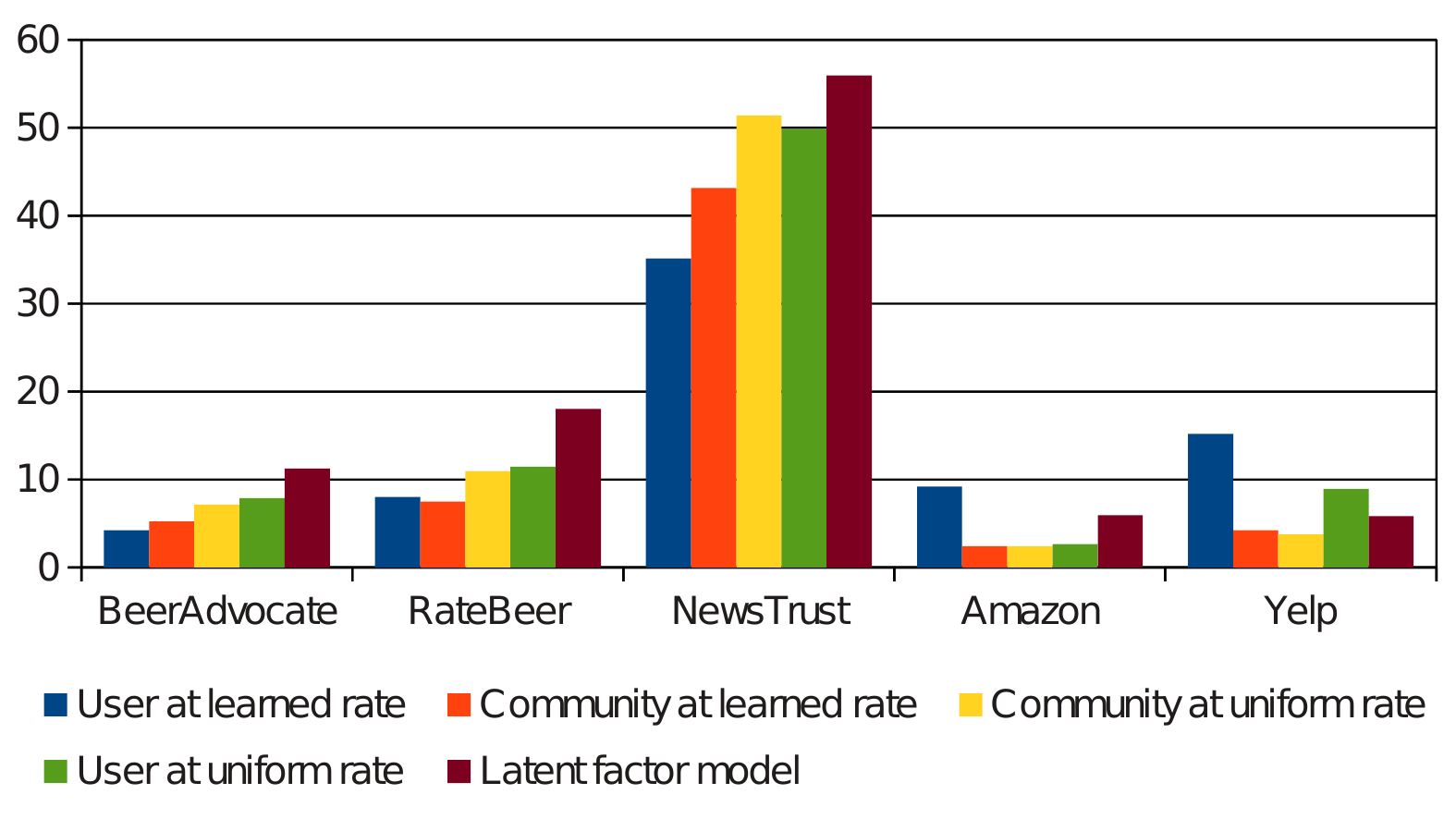}
 \caption{MSE improvement (\%) of our model over baselines.}
  \label{fig:improvement}
  \vspace{-1.5em}
\end{figure}

\noindent {\bf Baselines:} We consider the following baselines for our work, and use the available code\footnote{http://cseweb.ucsd.edu/~jmcauley/code/} for experimentation.
\squishlist
\item[a)]\emph{LFM}: A standard latent factor recommendation model~\cite{korenKDD2008}.
\item[b)]\emph{Community at uniform rate}: Users and products in a community evolve using a single ``global clock''~\cite{KorenKDD2010}\cite{xiongSDM2010}\cite{XiangKDD2010}, where the different stages of the community evolution appear at uniform time intervals. So the community prefers different products at different times.
\item[c)]\emph{Community at learned rate}: This extends (b) by learning the rate at which the community evolves with time, eliminating the uniform rate assumption.
\item[d)]\emph{User at uniform rate}: This extends (b) to consider individual users, by modeling the different stages of a user's progression based on preferences and experience levels evolving over time. 
The model assumes a uniform rate for experience progression. 
\item[e)]\emph{User at learned rate}: This extends (d) by allowing each user to evolve on a ``personal clock'', so that the time to reach certain experience levels depends on the user \cite{mcauleyWWW2013}.
\squishend
   f) \emph{Our model with past experience level}: In order to determine how well our model captures {\em evolution of user experience over time},
we consider another baseline
where we
\emph{randomly sample} the experience level reached by users at some timepoint {\em previously} in their lifecycle, who may have evolved thereafter. 
We learn our model parameters from the data up to this time, and again predict the user's most recent three item ratings. Note that this baseline considers textual content of user contributed reviews, unlike other baselines that ignore them. Therefore it is better than vanilla content-based methods, with the notion of past evolution, and is the strongest baseline for our model.

\noindent {\bf Discussions:} Table~\ref{fig:MSE} compares the \emph{mean squared error (MSE)} for rating predictions, generated by our model versus the six baselines. Our model consistently outperforms all baselines, reducing the MSE by ca. $5$ to $35\%$. Improvements of our model over baselines are statistically significant at p-value $<0.0001$.

Our performance improvement is most prominent for the NewsTrust community, which exhibits strong language features, and topic polarities in reviews. The lowest improvement (over the best performing baseline in any dataset) is achieved for Amazon movie reviews. A possible reason is that the community is very diverse with a very wide range of movies and that review texts heavily mix statements about movie plots with the actual review aspects like praising or criticizing certain facets of a movie. The situation is similar for the food and restaurants case. Nevertheless, our model always wins over the best baseline from {\em other} works, which is typically the ``user at learned rate" model.

\noindent {\bf Evolution effects:} We observe in Table~\ref{fig:MSE} that our model's predictions degrade when applied to the users' {\em past} experience level, compared to their {\em most recent} level. This signals that the model captures user evolution past the previous timepoint. Therefore the last (i.e., most recent) experience level attained by a user is most informative for generating new recommendations.

\subsection{Qualitative Analysis}

\noindent{\bf Salient words for facets and experience levels}: We point out typical word clusters, with {\em illustrative} labels, to show the variation of language for users of different experience levels and different facets. Tables \ref{tab:amazonTopics} and \ref{tab:beerTopics} show salient words to describe the beer facet {\em taste} and movie facets {\em plot} and {\em narrative style}, respectively -- at different experience levels. Note that the facets being latent, their labels are merely our interpretation. Other similar examples can be found in Tables~\ref{tab:facetWords} and~\ref{tab:newstrustTopics}.

BeerAdvocate and RateBeer are very focused communities; so it is easier for our model to characterize the user experience evolution by vocabulary and writing style in user reviews. We observe in Table~\ref{tab:beerTopics} that users write more descriptive and {\em fruity} words to depict the beer taste as they become more experienced.

For movies, the wording in reviews is much more diverse and harder to track.
Especially for blockbuster movies, which tend to dominate this data, the reviews mix all kinds of aspects.
A better approach here could be to focus on specific kinds of movies (e.g., by genre or production studios) that may better distinguish experienced users from amateurs or novices in terms of their refined taste and writing style.

\begin{table}
\centering
\small
\begin{tabular}{p{8cm}}
\toprule
\textbf{Experience Level 1:} drank, bad, maybe, terrible, dull, shit
\\\midrule
\textbf{Experience Level 2:} bottle, sweet, nice hops, bitter, strong light, head, smooth, good, brew, better, good
\\\midrule
\textbf{Expertise Level 3:} sweet alcohol, palate down, thin glass, malts, poured thick, pleasant hint, bitterness, copper hard
\\\midrule
\textbf{Experience Level 4:} smells sweet, thin bitter, fresh hint, honey end, sticky yellow, slight bit good, faint bitter beer, red brown, good malty, deep smooth bubbly, damn weak
\\\midrule
\textbf{Experience Level 5:} golden head lacing, floral dark fruits, citrus sweet, light spice, hops, caramel finish, acquired taste, hazy body, lacing chocolate, coffee roasted vanilla, creamy bitterness, copper malts, spicy honey
\\
\bottomrule
\end{tabular}
\caption{Experience-based facet words for the {\em illustrative} beer facet {\em taste}.}
\label{tab:beerTopics}
\vspace{-1.5em}
\end{table}

\noindent{\bf MSE for different experience levels}: We observe a weak trend that the MSE decreases with increasing experience level. Users at the highest level of experience almost always exhibit the lowest MSE. So we tend to better predict the rating behavior for the most mature users than for the remaining user population. This in turn enables generating better recommendations for the ``connoisseurs" in the community.

\noindent{\bf Experience progression}: Figure~\ref{fig:progression} shows the proportion of reviews written by community members at different experience levels right before advancing to the next level. Here we plot users with a minimum of $50$ reviews, so they are
certainly not ``amateurs".
A large part of the community progresses from level $1$ to level $2$. However, from here only few users move to higher levels, leading to a skewed distribution. We observe that the majority of the population stays at level $2$.




\begin{figure}
\centering
 \includegraphics[scale=0.5]{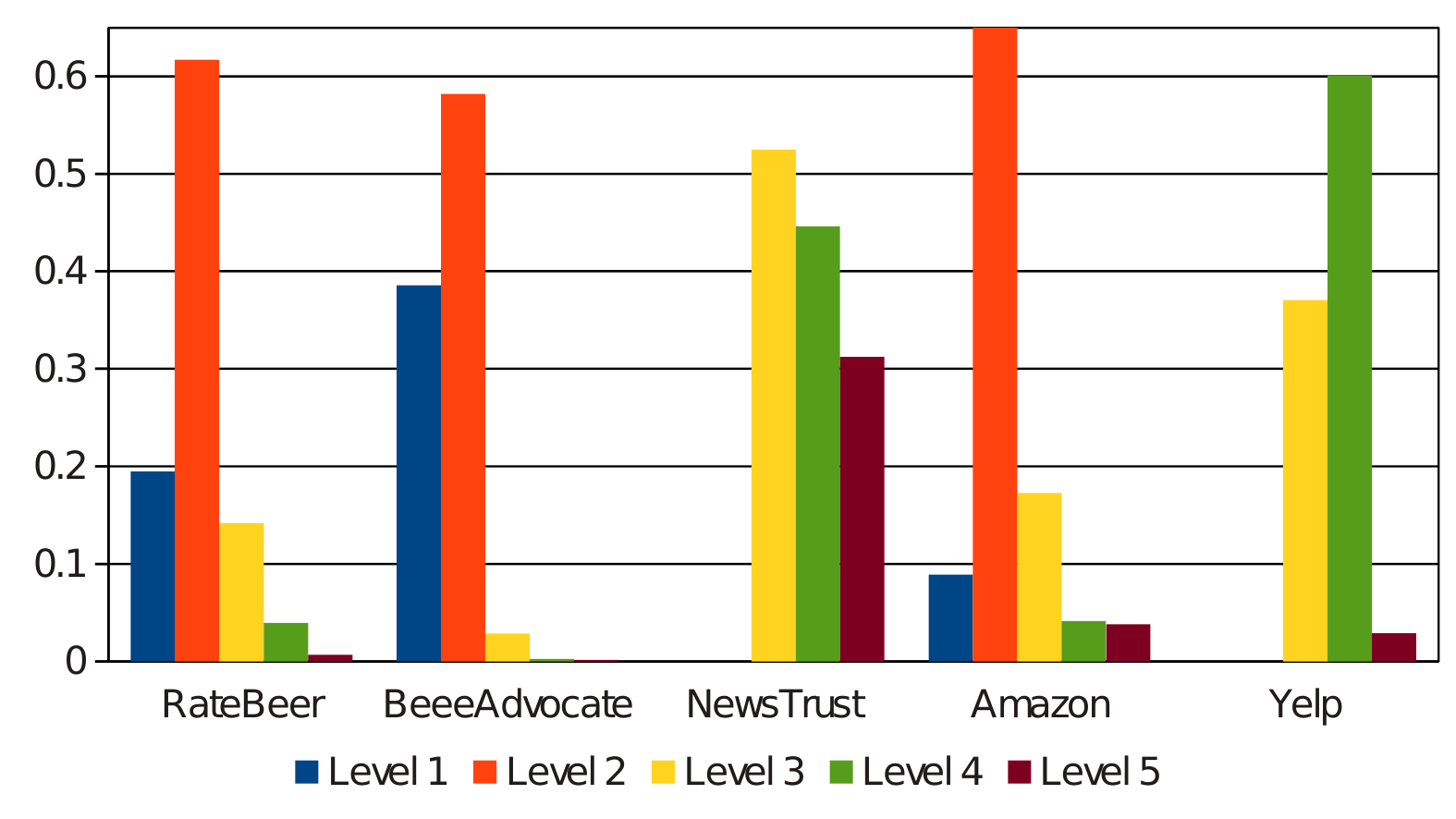}
   \vspace{-.5em}
 \caption{Proportion of reviews at each experience level of users.}
  \label{fig:progression}
  \vspace{-1.5em}
\end{figure}

\noindent{\bf User experience distribution}:
Table~\ref{tab:userExperienceDistr} shows the number of users per experience level in each domain, for users with $>50$ reviews. The distribution also follows our intuition of a highly skewed distribution.
Note that almost all users with $<50$ reviews belong to levels 1 or 2.

\begin{table}[t]
\centering
 \begin{tabular}{lccccc}
 \toprule
\bf{Datasets} &	\bf{e=1} & \bf{e=2} & \bf{e=3} & \bf{e=4} &\bf{e=5}\\
\midrule
BeerAdvocate & 0.05 & 0.59 & 0.19 & 0.10 & 0.07\\
RateBeer & 0.03 & 0.42 & 0.35 & 0.18 & 0.02\\
NewsTrust & - & - & 0.15 & 0.60 & 0.25\\
Amazon & - & 0.72 & 0.13 & 0.10 & 0.05\\
Yelp & - & - & 0.30 & 0.68 & 0.02\\
\bottomrule
 \end{tabular}
 \caption{\small Distribution of users at different experience levels.}
 \label{tab:userExperienceDistr}
 \vspace{-1em}
\end{table}

\noindent{\bf Language model and facet preference divergence}: Figure~\ref{fig:modelFacet} and~\ref{fig:modelLang} show the $KL$ divergence for facet-preference and language models of users at different experience levels, as computed by our model.
The facet-preference divergence increases with the gap between experience levels, but not as {\em smooth} and prominent as for the language models. 
On one hand, this is due to the complexity of {\em latent} facets vs. {\em explicit} words.
On the other hand, this also affirms our notion of grounding the model on \emph{language}. 

\begin{figure*}
    \centering
    \begin{subfigure}[b]{\textwidth}
        \centering
        \includegraphics[width=0.8\columnwidth]{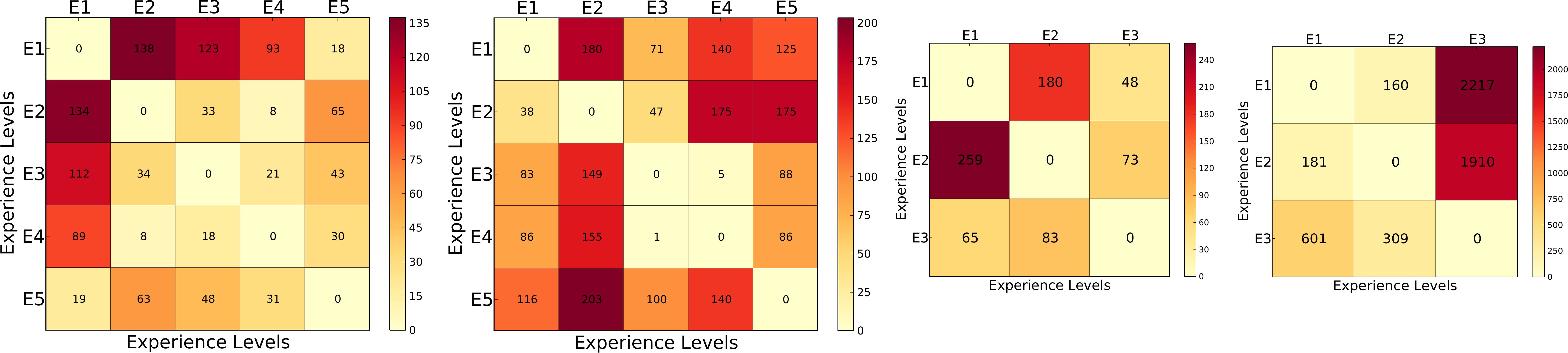}
        \caption{User at learned rate~\cite{mcauleyWWW2013}: Facet preference divergence with experience.}
        \label{fig:baselineFacet}
    \end{subfigure}
    \begin{subfigure}[b]{\textwidth}
        \centering
        \includegraphics[width=0.8\columnwidth]{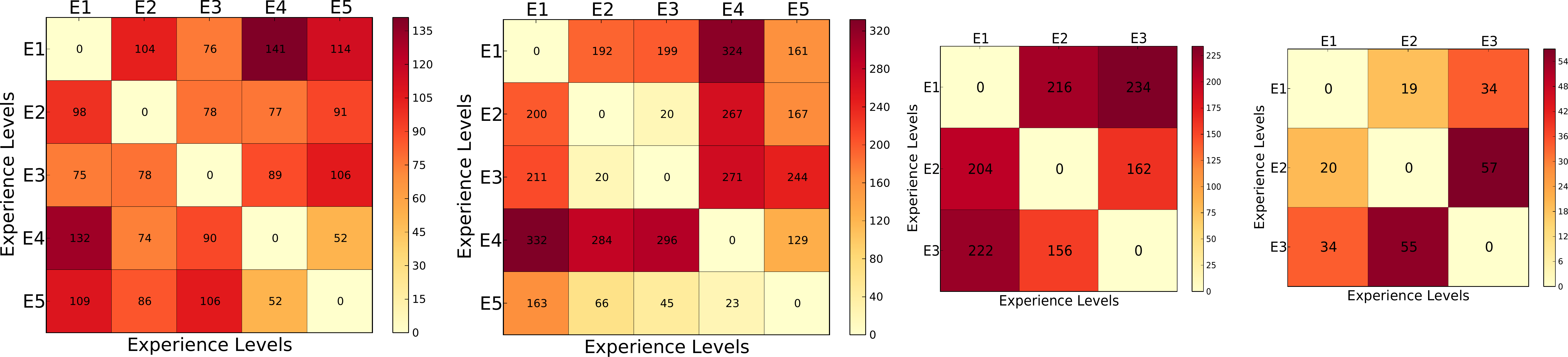}
        \caption{Our model: Facet preference divergence with experience.}
        \label{fig:modelFacet}
    \end{subfigure}
    \begin{subfigure}[b]{\textwidth}
        \centering
        \includegraphics[width=0.8\columnwidth]{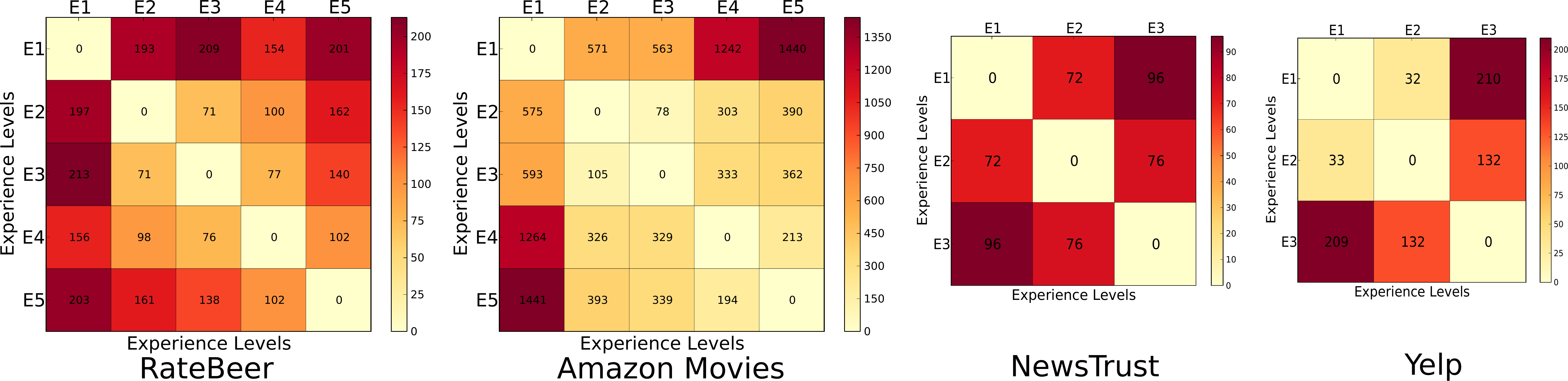}
        \caption{Our model: Language model divergence with experience.}
        \label{fig:modelLang}
    \end{subfigure}
    \caption{Facet preference and language model $KL$ divergence with experience.}
    \label{fig:heatmaps}
    \vspace{-1.5em}
\end{figure*}

\noindent{\bf Baseline model divergence}: Figure~\ref{fig:baselineFacet} shows the facet-preference divergence of users at different experience levels computed by the baseline model ``user at learned rate''~\cite{mcauleyWWW2013}.
The contrast between the heatmaps of our model and the baseline is revealing. The increase in divergence with increasing gap between experience levels is very {\em rough} in the baseline model, although the trend is obvious. 

\section{Use-Case Study}
\label{sec:usecases}

So far we have focused on traditional item recommendation for items like beers or movies. Now we switch to a different kind of items - newspapers and news articles - tapping into the NewsTrust online community ({\tt \small newstrust.net}). NewsTrust features news stories posted and reviewed by members, 
many of whom are professional journalists and content experts. Stories are reviewed based on their objectivity, rationality, and general quality of language to present an unbiased and balanced narrative of an event. The focus is on \emph{quality journalism}.

In our framework, each story is an item, which is rated and reviewed by a user. The facets are the underlying topic distribution of reviews, with topics being {\em Healthcare, Obama Administration, NSA}, etc. The facet preferences can be mapped to the (political) polarity of users in the news community. 

\noindent {\bf Recommending News Articles:}
Our first objective is to recommend news to readers catering to their facet preferences, viewpoints, and experience. We apply our joint model to this task, and compare the predicted ratings with the ones observed for withheld reviews in the NewsTrust community. The mean squared error (MSE) results are reported in Table~\ref{fig:MSE} in Section~\ref{sec:experiments}. Table~\ref{tab:newstrustTopics} shows salient examples of the vocabulary by users at different experience levels on the topic {\em US Election}.

\begin{table}
\small
\begin{tabular}{p{8cm}}
\toprule
\textbf{Level 1:} bad god religion iraq responsibility\\
\textbf{Level 2:} national reform live krugman questions clear jon led meaningful lives california powerful safety impacts\\
\textbf{Level 3:} health actions cuts medicare nov news points oil climate major jobs house high vote congressional spending unemployment strong taxes citizens events failure\\
\bottomrule
\end{tabular}
\caption{Salient words for the {\em illustrative} NewsTrust topic {\em US Election} at different experience levels.}
\label{tab:newstrustTopics}
\vspace{-1em}
\end{table}


\begin{table}[t]
\begin{center}
 \begin{tabular}{lcc}
 \toprule
\bf{Models} & \bf{$F_1$} & \bf{$NDCG$}\\
\midrule
User at learned rate~\cite{mcauleyWWW2013} & 0.68 & 0.90 \\
Our model & 0.75 & 0.97 \\
\bottomrule
 \end{tabular}
\caption{Performance on identifying experienced users.}
\label{tab:ndcgUsers}
\end{center}
\vspace{-3em}
\end{table}

\noindent {\bf Identifying Experienced Users:}
Our second task is to find experienced members of this community, who have potential for being \emph{citizen journalists}. In order to find how good our model predicts the experience level of users, we consider the following as ground-truth for user experience. In NewsTrust, users have {\em Member Levels} calculated by the NewsTrust staff based on  community engagement, time in the community, other users' feedback on reviews, profile transparency, and manual validation.
We use these member levels to categorize users as {\em experienced} or {\em inexperienced}. This is treated as the ground truth for assessing the prediction and ranking quality of our model and the baseline ``user at learned rate'' model~\cite{mcauleyWWW2013}.
Table~\ref{tab:ndcgUsers} shows the $F_1$ scores of these two competitors.
We also computed the \emph{Normalized Discounted Cumulative Gain (NDCG)} \cite{Jarvelin:TOIS2002} 
for the ranked lists of users generated by the two models. NDCG gives geometrically decreasing weights to predictions at various positions of ranked list:
{\small
$NDCG_p = \frac{DCG_p}{IDCG_p}$ where
$DCG_p = rel_1 + \sum_{i=2}^p \frac{rel_i}{\log_2 i}$
}

Here, $rel_i$ is the relevance ($0$ or $1$) of a result at position $i$. 

As Table~\ref{tab:ndcgUsers} shows, our model clearly outperforms the baseline model on both $F_1$ and $NDCG$.

\section{Related Work}

State-of-the-art recommenders based on collaborative filtering \cite{korenKDD2008}\cite{koren2011advances} exploit user-user and item-item similarities by latent factors. Explicit user-user interactions have been exploited in trust-aware recommendation systems~\cite{GuhaWWW2004}\cite{West-etal:2014}. The temporal aspects leading to bursts in item popularity, bias in ratings, or the evolution of the entire community as a whole is studied in~\cite{KorenKDD2010}\cite{xiongSDM2010}\cite{XiangKDD2010}. Other papers have studied temporal issues for anomaly detection~\cite{Gunnemann2014}, detecting changes in the social neighborhood~\cite{MaWSDM2011} and linguistic norms~\cite{DanescuWWW2013}. However, none of these prior work has considered the evolving experience and behavior of individual users.

The recent work\cite{mcauleyWWW2013}, which is one of our baselines, modeled the influence of rating behavior on evolving user experience. However, it ignores the vocabulary and writing style of users in reviews, and their natural {\em smooth} temporal progression. In contrast, our work considers the review texts for additional insight into facet preferences and {\em smooth} experience progression.

Prior work that tapped user review texts focused on other issues. Sentiment analysis over reviews aimed to learn latent topics~\cite{linCIKM2009}, latent aspects and their ratings~\cite{lakkarajuSDM2011}\cite{wang2011}, and user-user interactions~\cite{West-etal:2014}. \cite{mcauleyRecSys2013}\cite{WangKDD2011} unified various approaches to generate user-specific ratings of reviews. \cite{mukherjeeSDM2014} further leveraged the author writing style. However, all of these prior approaches operate in a static, snapshot-oriented manner, without considering time at all.

From the modeling perspective, some approaches learn a document-specific discrete rating~\cite{linCIKM2009}\cite{ramageKDD2011}, whereas others learn the facet weights outside the topic model (e.g., ~\cite{lakkarajuSDM2011, mcauleyRecSys2013, mukherjeeSDM2014}). In order to incorporate continuous ratings,~\cite{bleiNIPS2007} proposed a complex and computationally expensive Variational Inference algorithm, and~\cite{mimnoUAI2008} developed a simpler approach using Multinomial-Dirichlet Regression. The latter inspired our technique for incorporating supervision.

A general (continuous) version of this work is presented in~\cite{DBLP:conf/kdd/MukherjeeGW16} with fine-grained temporal evolution of user experience, and resulting language model using Geometric Brownian Motion and Brownian Motion, respectively.

\section{Conclusion}

Current recommender systems do not consider user experience when generating recommendations. 
In this paper, 
we have proposed an experience-aware recommendation model that can adapt to the changing preferences and maturity of users in a community.
We model the {\em personal evolution} of a user in rating items that she will appreciate at her current maturity level. 
We exploit the coupling between the \emph{facet preferences} of a user, her \emph{experience}, \emph{writing style} in reviews, and \emph{rating behavior} to capture the user's temporal evolution.
Our model is the first work that considers the progression of user experience as expressed in the text of item reviews.


Our experiments -- with data from domains like beer, movies, food, and news -- demonstrate that our model substantially reduces the mean squared error for predicted ratings, compared to the state-of-the-art baselines. 
This shows our method can generate better recommendations than those models.
We further demonstrate the utility of our method in a use-case study about identifying experienced members in the NewsTrust community who can be potential citizen journalists.


\bibliographystyle{abbrv}
\bibliography{kdd14}

\end{document}